\documentclass{article}

\PassOptionsToPackage{numbers, compress}{natbib}

 \usepackage[main, final]{neurips_2026}

\usepackage[utf8]{inputenc} 
\usepackage[T1]{fontenc}    
\usepackage{hyperref}       
\usepackage{url}            
\usepackage{booktabs}       
\usepackage{amsfonts}       
\usepackage{nicefrac}       
\usepackage{microtype}      
\usepackage{xcolor}         

\usepackage{amsmath}
\usepackage{graphicx}
\usepackage{subcaption}
\usepackage{wrapfig}
\usepackage{xspace}
\usepackage{bm}
\usepackage{multirow}
\usepackage{makecell}
\usepackage{xcolor}
\usepackage{colortbl}
\newcommand{\ours}{ADTT\xspace}

\title{Aligned Delta-Triplane Transformers as\\Occupancy World Models}

\author{%
  Haoran Xu$^{1}$, Peixi Peng$^{1,3}$\thanks{Corresponding authors.},\hspace{0.5em}Shijun Guo$^{4}$, Guang Tan$^{2}$\footnotemark[1],\\
  \textbf{Yiqian Chang}$^{5}$, \textbf{Yisen Zhao}$^{1}$, \textbf{Shuaixian Wang}$^{6}$, and \textbf{Yonghong Tian}$^{1,3}$\\[0.5em]
  $^{1}$Peking University \quad $^{2}$Shenzhen Campus of Sun Yat-sen University \\
  $^{3}$Peng Cheng Laboratory \quad $^{4}$Xidian University \\
  $^{5}$Shenzhen Campus of Harbin Institute of Technology \\
  $^{6}$Shenzhen Polytechnic University \\ 
  \texttt{\{xuhr.sz, pxpeng\}@pku.edu.cn} \quad 
  \texttt{tanguang@mail.sysu.edu.cn} \\
}

\begin{document}

\maketitle

\begin{abstract}
Occupancy World Models (OWMs) aim to predict future 3D occupancy scenes from historical observations and future ego motions, providing a useful world simulation tool for autonomous driving. Existing methods usually rely on large networks to implicitly align multi-frame historical states and predict future occupancy in a full-state manner, which is costly and redundant because most scene regions remain unchanged over short time intervals. In this paper, we propose \emph{Aligned Delta-Triplane Transformer} (\ours), a compact 4D OWM that explicitly aligns historical triplanes with ego-motion compensation and predicts only future scene changes. Based on the aligned triplane prior, a query-conditioned Transformer predicts residual changes, where ego-motion queries guide controllable future motion and learnable external queries capture scene changes beyond ego motion. The predicted changes are added to the aligned prior and decoded into future occupancy scenes. Experiments on Occ3D-nus show that \ours achieves state-of-the-art forecasting performance with fewer parameters and higher FPS. Our code is publicly available online: \url{https://github.com/kyoran/ADTT}.
\end{abstract}

\begin{wrapfigure}[16]{r}{0.5\textwidth}
    \vspace{-27pt}
    \includegraphics[width=\linewidth]{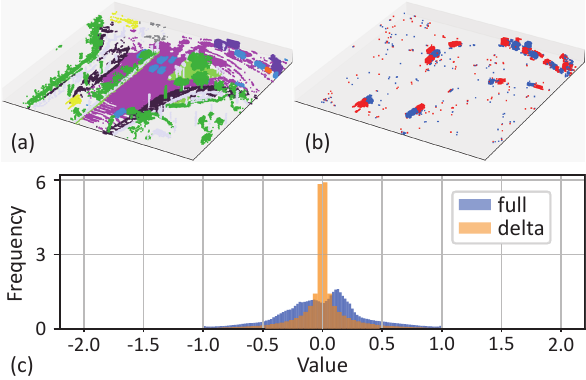}
    \vspace{-14pt}
    \caption{(a-b) Comparison between the full occupancy and occupancy changes ($\Delta$) of a scene. In the change map, red indicates newly appeared voxels, while blue denotes disappeared voxels. (c) Distribution of latent-space feature values for full occupancy versus occupancy changes.}
    \label{fig:moti}
\end{wrapfigure}
\section{Introduction}

Recently, Occupancy World Models (OWMs)~\cite{occworld,occsora,occllama,occllm,dome,uniscene,come,I2world,geniedrive} have emerged as a promising framework for modeling dynamic 3D scenes. Instead of representing the environment with 2D images~\cite{wang2024drivedreamer} or bird's-eye-view (BEV) features~\cite{li2025end}, OWMs describe real-world scenes with structured 3D semantic voxels and predict how these scenes evolve over time under different ego actions. This makes OWMs a useful tool for simulating future physical worlds, and provides a scalable way to generate training data, including rare corner cases, for autonomous driving and indoor navigation.

In designing OWMs, there are two tasks need to be solved: 
(i) removing historical ego-motion effects to better model surrounding object motion; and
(ii) using future ego motions to generate controllable future occupancy scenes.

For the first task, the observed occupancy changes usually come from two sources: ego-motion and the motion of surrounding objects. If the scene variation introduced by ego-motion can be removed, the remaining changes mainly come from moving objects, making future prediction much easier. This process is usually referred to as motion compensation. Most existing OWMs perform this step implicitly. They first compress high-dimensional 3D occupancy data into latent xy-plane features, and then rely on neural networks to learn the alignment across multiple frames. Such an implicit formulation often requires large model capacity to fit the geometric transformation between frames. Recent works such as $I^2$-World~\cite{I2world} and GenieDrive~\cite{geniedrive} introduce dedicated ego-motion-related attention mechanisms to ease the learning of temporal correspondence. However, their alignment is mainly learned through network attention rather than explicit 3D geometric transformation.

\begin{figure}[!t]
\centering
\includegraphics[width=\linewidth]{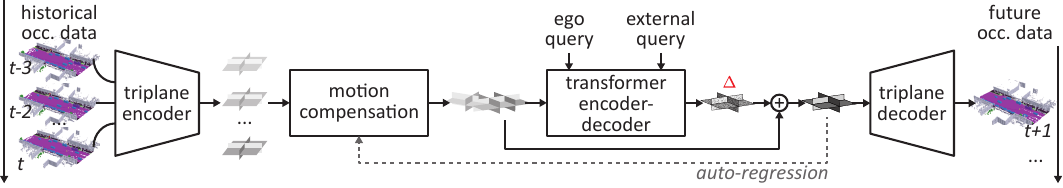}
\caption{\ours follows a simple yet effective procedure: it first represents occupancy scenes with pretrained triplanes, then explicitly aligns historical triplanes through historical ego-motion compensation, and finally employs a Transformer to predict triplane changes ($\Delta$) based on the aligned triplane features, future ego-motion query, and external learnable query.}
\label{fig:moti_ours}
\end{figure}

We provide an intuitive example in \figurename~\ref{fig:moti}. \figurename~\ref{fig:moti}(a) shows the full occupancy state in Occ3D-nus benchmark~\cite{nuscenes,occ3d}, while \figurename~\ref{fig:moti}(b) shows the occupancy changes between two adjacent frames after compensating the previous frame to the next timestep. It is clear that occupancy changes are inherently sparse and key elements of interest are more distinct in the change map. We further compare their latent-space distributions in \figurename~\ref{fig:moti}(c), where the $x$ axis denotes a scalar feature value. Full occupancy states exhibit a scattered distribution, while occupancy change states are much more tightly concentrated around zero, which reduces variance and is easier to learn.

For the second task, future ego-motion should be used as a condition to guide scene prediction. Existing OWMs typically feed ego-motion signals into large Transformers or diffusion models to directly predict the full future occupancy state~\cite{occllm,occllama}. However, as suggested by \figurename~\ref{fig:moti}, full-state prediction is often redundant, since most regions remain unchanged after motion compensation. This has motivated a compensation-then-prediction paradigm, where the model first removes ego-motion effects and then predicts future occupancy changes. Although this reduces the prediction burden, current methods~\cite{I2world,geniedrive} mainly use future ego-motion as the only condition for prediction. Such signals specify how the ego coordinate frame moves and which areas may enter the future view, but they do not provide enough guidance for the uncertain content that should appear in the future scene. As a result, the predicted future scenes may have limited temporal consistency.

Based on these observations, we propose \emph{Aligned Delta-Triplane Transformer} (\ours), a compact 4D OWM that explicitly aligns historical triplanes and predicts sparse future changes in the triplane latent space, as shown in \figurename~\ref{fig:moti_ours}. Specifically, we first pretrain a triplane encoder-decoder to obtain compact and geometry-aware occupancy representations. Then, using ego poses recorded at different timesteps, we perform ego-motion compensation to warp historical triplanes into the target prediction frame. This gives the predictor a stable geometric prior, so that it only needs to model residual changes that cannot be explained by ego motion. To predict these residuals, we design a query-conditioned Transformer. Ego-motion queries inject deterministic future motion guidance, while learnable external queries provide additional flexibility for modeling scene changes that are not explicitly controlled by ego motion. The predicted delta triplanes are added to the motion-aligned prior and decoded into future occupancy scenes. In summary, our contributions are three-fold:


$\bullet$ We propose \ours, a compact 4D autoregressive OWM that aligns historical triplanes with ego-motion compensation and forecasts sparse future scene changes.

$\bullet$ Within \ours, we introduce two key modules: triplane motion compensation, which aligns historical triplanes into the target frame, and query-conditioned delta prediction, which predicts residual changes from the aligned triplane using ego-motion queries for controllable future guidance and learnable external queries for scene dynamics beyond ego motion.

$\bullet$ Extensive experiments on Occ3D-nus~\cite{nuscenes,occ3d} validate the state-of-the-art (SOTA) performance of \ours in short- and long-term forecasting, cross-dataset generalization, and real-time inference.

\section{Related Works}

\textbf{3D occupancy reconstruction.} 
3D occupancy reconstruction methods~\citep{occnet,occ3d,surroundocc,liu2024volumetric,marinello2025camera} aim to represent the environment as a 3D voxel grid where each voxel encodes both geometric and semantic information. The field was pioneered by SSCNet~\citep{song2017semantic} for indoor scenes using depth sensors, and later extended to outdoor camera-based settings by MonoScene~\citep{cao2022monoscene}. These approaches primarily utilize multi-camera RGB images~\citep{li2023voxformer,huang2023tri} or LiDAR point clouds~\citep{cao2024pasco,xia2023scpnet} to infer voxel-wise occupancy and semantics in an agent-centric coordinate system. A core challenge lies in building accurate correlations between raw sensor data and 3D voxel space. Once an accurate occupancy reconstruction is achieved, further learning of the temporal dynamics of scene changes becomes necessary~\citep{viewformer}.

\textbf{4D occupancy prediction.} 
To anticipate future scene changes, various 4D occupancy prediction methods have been proposed to capture the temporal dynamics of scene evolution. Some~\citep{lu2021monet,mersch2022self,khurana2023point} predict future sensor-level data, which is subsequently voxelized into occupancy results, while others~\citep{cam4docc,occprophet,xu2024spatiotemporal,cmop} directly forecast occupancy outcomes from historical observations. These methods mainly aim to reduce spatio-temporal prediction errors, but they generally do not consider controllable forecasting conditioned on ego motion.

\textbf{World models for autonomous driving.} World models~\citep{ha2018world,ha2018recurrent} aim to compress high-dimensional scene representations to capture the temporal dynamics of scene transitions, facilitating both future scene predictions and motion planning for the agent. In autonomous driving, existing models~\citep{vad,zhang2024bevworld} typically map surrounding traffic participants to the BEV perspective to predict instance-level tracklets or directly use diffusion models~\citep{wu2024holodrive,wang2024driving,wang2024drivedreamer,gao2025vista,jia2023adriver} to generate pixel-level future driving views. Recent world models~\citep{occllama,occllm,occworld,renderworld,dome} further exploit structured 3D occupancy data for future scene prediction. However, most of them still compress occupancy scenes into BEV latent features and rely on Transformers or diffusion models to implicitly align multi-frame historical context, which often requires large model capacity to forecast the full future scene. In contrast, our method explicitly aligns historical features and predicts sparse scene changes, both in the triplane latent space~\cite{geniedrive,huang2023tri}.

\section{Methodology}

\subsection{Problem Formulation}

Let $S^t\in\mathbb{R}^{H\times W\times L}$ denote the semantic occupancy scene at timestep $t$ in the ego-centric coordinate frame, where $H$, $W$, and $L$ denote the height, width, and length of the voxel space. Given a sequence of historical occupancy observations and ego motions, the model uses them as predefined scene context and predicts future occupancy scenes under given future ego motions.

For clarity, we denote the historical occupancy context as
\begin{equation}
    \mathcal{S}_{\rm hist}^{t}
    =
    \{S^{t-\tau_p+1},\cdots,S^{t}\},
\end{equation}
and the ego motions associated with the observed and future intervals as
\begin{equation}
\mathcal{M}_{\rm hist}^{t}
=
\{m^{t-\tau_p+2},\cdots,m^{t}\},
\quad
\mathcal{M}_{\rm fut}^{t}
=
\{m^{t+1},\cdots,m^{t+\tau_f}\}.
\end{equation}

Here, each ego motion $m^i$ describes the ego transformation from timestep $i-1$ to timestep $i$, such as translation and rotation. The historical occupancy sequence provides scene context, including static structures, visible objects, and recent scene evolution. The historical ego motions are used to align observations captured under different ego-centric frames, while future ego motions serve as conditions for controllable future generation.

Formally, the OWM learns a mapping:
\begin{equation}
    \hat{S}^{t+1:t+\tau_f}
    =
    \bm{\Phi}
    \left(
    \mathcal{S}_{\rm hist}^{t},
    \mathcal{M}_{\rm hist}^{t},
    \mathcal{M}_{\rm fut}^{t}
    \right),
    \label{eq:problem}
\end{equation}
where $\hat{S}^{t+1:t+\tau_f}$ denotes the predicted future occupancy sequence.

\begin{wrapfigure}[22]{r}[0cm]{0px}
\begin{minipage}{7.8cm}
    \centering
    \includegraphics[width=\linewidth]{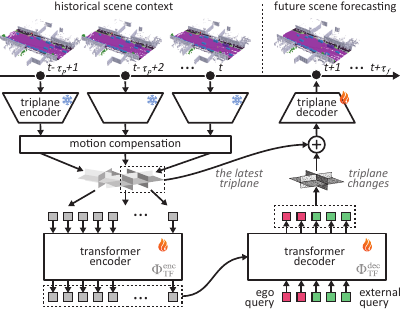}
    \vspace{-12pt}
    \caption{\ours performs motion compensation and tokenization in the triplane latent space. The Transformer encoder then encodes historical context from triplane tokens, while the decoder predicts next-frame triplane changes using deterministic ego-motion queries and learnable external queries. These predicted changes are added to the latest triplane and finally decoded into future occupancy states.}
    \label{fig:meth:framework}
\end{minipage}
\end{wrapfigure}
Accordingly, as shown in \figurename~\ref{fig:meth:framework}, we propose \emph{Aligned Delta-Triplane Transformer} (\ours), which aligns historical states and predicts sparse future changes in the aligned triplane latent space.


\subsection{Ego-Motion-Aligned Triplane Representation}

Recent occupancy-related works~\cite{geniedrive,huang2023tri} have gradually shifted their latent space from BEV features to triplane representations, since triplanes provide a compact latent space while preserving more explicit spatial structure. Following this trend, we build our OWM on a pretrained triplane latent space. Different from prior methods that mainly use triplanes as compact scene encodings, we further exploit their geometric structure for ego-motion compensation, explicitly aligning historical triplanes across ego-centric frames before predicting sparse future changes.

\begin{wrapfigure}[18]{r}[0cm]{0px}
\begin{minipage}{2.81cm}
    \centering
    \vspace{-13pt}
    \includegraphics[width=0.98\linewidth]{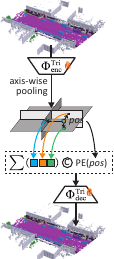}
    \vspace{-6mm}
    \caption{Pretrained triplane for OWM.}
    \label{fig:meth:enc_dec}
\end{minipage}
\end{wrapfigure}

As shown in \figurename~\ref{fig:meth:enc_dec}, the occupancy scene $S^t$ is encoded by $\mathbf{\Phi}_{\rm Tri}^{\rm enc}$ and compressed into three orthogonal feature planes:
\begin{equation}
    s^t = \{s^t_{xy}, s^t_{xz}, s^t_{yz}\},
\end{equation}
where
\begin{equation}
    s^t_{xy}\in\mathbb{R}^{C\times H_{xy}\times W_{xy}},\quad
    s^t_{xz}\in\mathbb{R}^{C\times H_{xz}\times W_{xz}},\quad
    s^t_{yz}\in\mathbb{R}^{C\times H_{yz}\times W_{yz}}.
\end{equation}

The three planes provide complementary projections of the 3D scene. To reconstruct the occupancy scene, each 3D point $pos=(x,y,z)$ queries its corresponding features from the $xy$, $xz$, and $yz$ planes. The queried features are fused with positional encoding ${\rm PE}(pos)$ and passed into $\mathbf{\Phi}_{\rm Tri}^{\rm dec}$ to predict the semantic occupancy label.

We pretrain $\mathbf{\Phi}_{\rm Tri}^{\rm enc}$ and $\mathbf{\Phi}_{\rm Tri}^{\rm dec}$ with occupancy reconstruction:
\begin{equation}
J_{\rm enc,dec}
=
\mathbb{E}_{t\sim \mathcal{T}}
\left[
    \mathcal{L}_{occ}
    \left(
    \mathbf{\Phi}_{\rm Tri}^{\rm dec}(\mathbf{\Phi}_{\rm Tri}^{\rm enc}(S^t)), S^t
    \right)
\right],
\label{eq:loss_enc_dec}
\end{equation}
where $\mathcal{T}$ denotes the set of timesteps in the training data, and
$\mathcal{L}_{occ}=\mathcal{L}_{ce}+\lambda\mathcal{L}_{lz}$ combines cross-entropy and Lovasz-softmax losses~\citep{berman2018lovasz,semcity}. After pretraining, $\bm{\Phi}_{\rm enc}$ provides compact and geometry-aware triplane states for future prediction.

We then perform ego-motion compensation using the known ego poses at different timesteps. These poses describe the trajectory of the ego vehicle and are recorded by the vehicle localization system, such as odometry and IMU~\cite{nuscenes}. Inspired by spatial transformation and differentiable warping~\cite{jaderberg2015spatial,zhou2017unsupervised}, we design an ego-motion compensation operation in the triplane latent space.

For each timestep $k$, we denote its ego pose in the global coordinate frame as $(\mathbf{R}_k,\mathbf{T}_k)$, where $\mathbf{R}_k$ and $\mathbf{T}_k$ transform a point from the ego coordinate frame at timestep $k$ to the global coordinate frame:
\begin{equation}
    \mathbf{p}_{\rm global}^{k}
    =
    \mathbf{R}_k \mathbf{p}_{\rm local}^{k}
    +
    \mathbf{T}_k .
    \label{eq:ego_to_global}
\end{equation}
Here, $\mathbf{p}_{\rm local}^{k}$ denotes a point in the ego coordinate frame at timestep $k$, and $\mathbf{p}_{\rm global}^{k}$ denotes the same point in the global coordinate frame.

Given a historical timestep $i$ and a target prediction timestep $j$, our goal is to align the historical triplane $s^i$ to the target ego coordinate frame at timestep $j$. For each location on the target plane, we first regard it as a 3D point $\mathbf{p}_{\rm local}^{j}$ in the target ego coordinate frame, and then find where this point should sample from in the historical triplane. Specifically, we first map $\mathbf{p}_{\rm local}^{j}$ to the global coordinate frame using the target ego pose, and then map it back to the historical ego coordinate frame:
\begin{equation}
    \mathbf{p}_{\rm local}^{i}
    =
    \mathbf{R}_i^\top
    \left(
    \mathbf{R}_j \mathbf{p}_{\rm local}^{j}
    +
    \mathbf{T}_j
    -
    \mathbf{T}_i
    \right).
    \label{eq:target_to_source}
\end{equation}

\begin{wrapfigure}[11]{r}[0cm]{0px}
\begin{minipage}{6.5cm}
    \centering
    \includegraphics[width=0.98\linewidth]{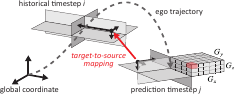}
    \caption{Motion compensation in the triplane latent space.}
    \label{fig:meth:motion_comp}
\end{minipage}
\end{wrapfigure}
This transformation gives the corresponding source-frame location for each target-frame point. In practice, if the projected location lies inside the valid range of the historical plane, we copy the nearest source feature to the target frame. Otherwise, the location is filled with zeros, indicating that this region has no valid historical correspondence and should be inferred by the prediction module.

We denote the motion-compensated triplane as:
\begin{equation}
    \overline{s}^{j}
    =
    \{\overline{s}_{xy}^{j}, \overline{s}_{xz}^{j}, \overline{s}_{yz}^{j}\},
\end{equation}
which provides an aligned geometric prior for predicting the target scene at timestep $j$.

\subsection{Query-Conditioned Delta Prediction}

Given the motion-aligned triplane representation, \ours predicts the triplane changes $\Delta s_{t+1}$ with a Transformer encoder-decoder~\cite{vaswani2017attention,carion2020end}. The encoder builds a spatio-temporal memory from historical triplanes, while the decoder predicts future state changes using ego queries and external queries.

\textbf{Triplane tokenization.}
We first convert the motion-compensated triplane $\overline{s}^{j}$ into a low-resolution 3D latent token grid, allowing the Transformer to model the scene in a unified and compact spatial structure. Specifically, each plane is projected and downsampled by a lightweight 2D convolutional tokenizer, denoted as $\phi_{xy}$, $\phi_{xz}$, and $\phi_{yz}$, with $D_{\rm down}$ output channels. As shown in \figurename~\ref{fig:meth:motion_comp}, the three plane features are then broadcast to the same 3D grid of size $G_x\times G_y\times G_z$ and fused. For each 3D grid location $pos$, the fused feature is computed as:
\begin{equation}
    v^{t}(pos)
    =
    {\rm Concat}_{uv\in\{xy, xz, yz\}}\left[
    {\rm Broad}_{uv}
    \left(
    \phi_{uv}(\overline{s}_{uv}^{t})
    \right)(pos)\right],
\end{equation}
where ${\rm Broad}_{uv}(\cdot)$ denotes broadcasting a plane feature to the 3D latent grid along its missing axis.

The fused feature is then projected to the Transformer token dimension and combined with 3D sinusoidal positional encoding:
\begin{equation}
    z^{j}(pos)
    =
    {\rm LN}
    \left(
    {\rm FC}(v^{j}(pos))
    \right)
    +
    {\rm PE}(pos),
    \label{eq:triplane_token}
\end{equation}
where ${\rm FC}(\cdot)$ denotes a fully connected layer that maps the feature to $D_{\rm token}$ dimensions, ${\rm LN}(\cdot)$ denotes layer normalization, and ${\rm PE}(pos)$ is the 3D spatial positional encoding.

\textbf{Historical memory encoding.}
For each prediction step, we use the most recent $\tau_p$ motion-compensated triplane states as temporal context. After tokenization, the historical token sequences are concatenated along the temporal dimension:
\begin{equation}
    Z_{\rm hist}^{t}
    =
    [z^{t-\tau_p+1};\cdots;z^{t}].
\end{equation}

We add a learnable temporal embedding to distinguish different history frames and feed the sequence into a Transformer encoder:
\begin{equation}
    \mathcal{M}^{t}
    =
    \mathbf{\Phi}_{\rm TF}^{\rm enc}
    \left(
    Z_{\rm hist}^{t}
    +
    e_{\rm time}
    \right),
    \label{eq:encoder_memory}
\end{equation}
where $\mathcal{M}^{t}$ denotes the encoded spatio-temporal memory. Through self-attention, the encoder allows tokens from different spatial locations and historical timesteps to interact, producing a memory bank that summarizes recent scene context.

\textbf{Ego-motion and external queries.}
The decoder queries serve as conditions for future scene prediction. Different from methods that only rely on ego motion, we construct the decoder query from two complementary sources. First, the future ego motion $m^{t+1}$ is encoded by an MLP and expanded into $N_{\rm ego}$ ego-motion queries:
\begin{equation}
    Q^{\rm ego}_{t+1}
    =
    f_{\rm ego}(m^{t+1})
    \in
    \mathbb{R}^{N_{\rm ego}\times D_{\rm token}}.
\end{equation}

These queries inject explicit information about the future ego transformation. Second, we introduce $N_{\rm ext}$ learnable external queries $Q^{\rm ext}\in\mathbb{R}^{N_{\rm ext}\times D_{\rm token}}$. Unlike ego-motion queries, these queries are not tied to a specific motion input. They provide additional learnable slots for modeling residual scene changes that are not fully determined by ego motion, such as local appearance changes, occlusion variations, and newly observed regions.

The final decoder query is obtained by concatenating the two query groups and adding a learnable future-step embedding:
\begin{equation}
    Q_{t+1}
    =
    [Q^{\rm ego}_{t+1}; Q^{\rm ext}]
    +
    e^{\rm step}_{t+1}.
    \label{eq:decoder_query}
\end{equation}

The total number of queries is set to match the number of target latent tokens, where $N=N_{\rm ego}+N_{\rm ext}=G_xG_yG_z$. The Transformer decoder attends to the encoded memory and predicts:
\begin{equation}
    \Delta Z^{t+1}
    =
    \mathbf{\Phi}_{\rm TF}^{\rm dec}
    \left(
    Q_{t+1}, \mathcal{M}^{t}
    \right),
    \label{eq:delta_tokens}
\end{equation}
where $\Delta Z^{t+1}\in\mathbb{R}^{N\times D_{\rm token}}$ denotes the predicted delta token sequence.

The predicted delta tokens are reshaped back to the predefined 3D latent grid and then upsampled $\text{Up}(\cdot)$ to the triplane resolution. We denote this operation as:
\begin{equation}
    \Delta s^{t+1}
    =
    {\rm Up}
    \left(
    {\rm Reshape}(\Delta Z^{t+1})
    \right),
    \label{eq:delta_upsample}
\end{equation}
where ${\rm Up}(\cdot)$ includes 3D feature refinement, interpolation, and projection back to the three triplane feature maps $\Delta s^{t+1} = \{\Delta s_{xy}^{t+1},\Delta s_{xz}^{t+1},\Delta s_{yz}^{t+1}\}$. The predicted delta triplane is added to the latest motion-aligned triplane prior to obtain the updated triplane:
\begin{equation}
    \hat{s}^{t+1}
    =
    \overline{s}^{t}
    +
    \Delta s^{t+1}.
    \label{eq:delta_update}
\end{equation}
Finally, the updated triplane is decoded into the future occupancy scene:
\begin{equation}
    \hat{S}^{t+1}
    =
    \mathbf{\Phi}_{\rm Tri}^{\rm dec}
    \left(
    \hat{s}^{t+1}
    \right).
\end{equation}

The future prediction module is trained with both latent-space and occupancy-space supervision:
\begin{equation}
J_{\rm fut}
=
\mathbb{E}_{t,k,uv}
\left[
    \mathcal{L}_{\rm tri}
    \left(
        \hat{s}^{uv}_{t+k}, s^{uv}_{t+k}
    \right)
    +
    \xi
    \mathcal{L}_{\rm occ}
    \left(
        \hat{S}^{t+k}, S^{t+k}
    \right)
\right],
\label{eq:loss_fut}
\end{equation}
where $\mathcal{L}_{\rm tri}$ is the sum of L1 and L2 regression losses, and $\xi$ balances triplane- and occupancy-level supervision. During training, we fix the pretrained encoder to keep the target triplanes stable and fine-tune the decoder to better reconstruct occupancy scenes from predicted latent states.

\section{Experiments}
\subsection{Experimental settings}

\textbf{Dataset and evaluation metrics.}
We evaluate occupancy forecasting on the widely used nuScenes dataset~\cite{nuscenes} adopted by OWM methods~\cite{occworld,occsora,occllama,occllm,dome,uniscene,come,I2world,geniedrive} with occupancy annotations from Occ3D~\cite{occ3d}, referred to as Occ3D-nus. The dataset contains 1,000 scenes, with 700 scenes for training and 100 scenes for testing. For the main evaluation, we use the past four frames (2 seconds) to predict the next six frames (3 seconds), which measures short-term forecasting performance. We further evaluate two types of zero-shot inference ability without additional training. First, following GenieDrive~\cite{geniedrive}, we directly evaluate the model trained on the 3-second setting for longer forecasting horizons. Second, following $I^2$-World~\cite{I2world}, we use Occ3D-Waymo~\cite{sun2020scalability,occ3d} to evaluate cross-dataset generalization on 202 validation scenes. For all tasks, we report intersection over union (IoU) for occupied and unoccupied voxels, as well as mean IoU (mIoU) over 18 semantic classes.

\textbf{Implementation details.}
At each timestep, the occupancy data $S^t$ has a resolution of $16\times200\times200$. It covers a spatial range of $[-40,40]$ meters along the X and Y axes and $[-1,5.4]$ meters along the Z axis, with a voxel size of $0.4$ meters. The pretrained triplane representation has a resolution of $16\times100\times100$ with 8 channels. During triplane tokenization, we use a 3D latent grid size of $(G_x,G_y,G_z)=(8,8,4)$, resulting in $N=G_x G_y G_z = 256$ latent tokens with token dimension $D_{\rm token}=96$. For each frame, the triplane tokens are first augmented with 3D sinusoidal positional encoding, and then with a learnable temporal embedding $e_{\rm time}$. In the Transformer decoder, we set the numbers of ego-motion queries and learnable external queries to $N_{\rm ego}=64$ and $N_{\rm ext}=192$, respectively. The upsampling operator $\text{Up}(\cdot)$ uses nearest-neighbor interpolation to recover the triplane resolution from the predicted latent tokens. Training is performed on 4 NVIDIA RTX A6000 GPUs, while testing is conducted on a single NVIDIA RTX 4090 GPU. Our code is available online\footnote{\url{https://github.com/kyoran/ADTT}}.

\begin{table}[!t]
\caption{Comparison on Occ3D-nus 4D occupancy forecasting at 1s, 2s, and 3s. The best and second-best results are shown in \textbf{bold} and \underline{underlined}, respectively.}
\label{tab:occ_compare_123}
\centering
\scriptsize
\begin{tabular}{
    r|c|
    ccc>{\columncolor{gray!20}}c|
    ccc>{\columncolor{gray!20}}c|
    cc
}
\toprule
    \multirow{2}{*}{Models} & 
    \multirow{2}{*}{Input} & 
    \multicolumn{4}{c|}{mIoU (\%)$\uparrow$} & 
    \multicolumn{4}{c}{IoU (\%)$\uparrow$} &
    \multirow{2}{*}{FPS$\uparrow$} &
    \multirow{2}{*}{Params$\downarrow$} \\
    & & 
    1s & 2s & 3s & Avg. & 
    1s & 2s & 3s & Avg. \\
\midrule
    OccWorld~\cite{occworld} & Occ &
        25.78 & 15.14 & 10.51 & 17.14 &
        34.63 & 25.07 & 20.18 & 26.63 & 
        18 & 72M \\
    OccSora~\cite{occsora} & Occ &
        32.77 & 22.04 & 14.40 & 23.07 &
        41.39 & 33.68 & 29.97 & 35.01 & 
        20 & 174M \\
    OccLLaMA~\cite{occllama} & {\tiny Occ\&Text} &
        25.05 & 19.49 & 15.26 & 19.93 &
        34.56 & 28.53 & 24.41 & 29.17 &
        - & > 7B \\
    OccLLM~\cite{occllm} & {\tiny Occ\&Text} &
        24.02 & 21.65 & 17.29 & 20.99 &
        36.65 & 32.14 & 28.77 & 32.52 &
        -  & > 7B \\
    DOME~\cite{dome} & Occ &
        35.11 & 25.89 & 20.29 & 27.10 &
        43.99 & 35.36 & 29.74 & 36.36 &
        6.54 & 397M \\
    UniScene~\cite{uniscene} & {\tiny Occ\&Box\&Map} &
        35.37 & 29.59 & 25.08 & 31.76 &
        38.34 & 32.70 & 29.09 & 34.84 &
        1.72 & 69M \\
    COME~\cite{come} & Occ &
        42.75 & 32.97 & 26.98 & 34.23 &
        50.57 & 43.47 & 38.36 & 44.13 & 
        0.30 & 692M \\
    $I^2$-World~\cite{I2world} & Occ & 
        47.62 & 38.58 & 32.98 & 39.73 & 
        54.29 & 49.43 & 45.69 & 49.80 &
        37 & 22M \\
    GenieDrive~\cite{geniedrive} & Occ &
        50.47 & 41.47 & 35.83 & 42.59 &
        56.87 & 51.46 & \textbf{47.08} & 51.80 &
        41 & 3.47M \\
    \rowcolor{gray!20}
    \ours (Ours) & Occ &
        \textbf{55.88} & \textbf{45.47} & \textbf{39.27} & \cellcolor{gray!50}\textbf{46.87} & 
        \textbf{59.15} & \textbf{51.89} & \underline{46.87} & \cellcolor{gray!50}\textbf{52.63} &
        \textbf{46} & \textbf{2.30M} \\
\midrule
    OccWorld-V & Camera & 
        10.97 & 15.16 & 10.19 & 7.53 & 
        20.68 & 24.81 & 20.09 & 17.14 & 
        3.77 & 132M \\
    DOME-V & Camera &
        17.79 & 14.23 & 11.58 & 14.53 &
        26.39 & 23.20 & 20.42 & 23.33 & 
        2.75 & 457M \\
    $I^2$-World-V & Camera &
        21.67 & 18.78 & 16.47 & 18.97 &
        30.55 & 28.76 & 26.99 & 28.77 &
        4.21 & 82M \\
    \rowcolor{gray!20}
    \ours-V (Ours) & Camera & 
        \textbf{24.15} & \textbf{20.37} & \textbf{17.63} & \cellcolor{gray!50}\textbf{20.71} &
        \textbf{33.62} & \textbf{31.05} & \textbf{28.49} & \cellcolor{gray!50}\textbf{31.05} & 
        \textbf{6.16} & \textbf{62.30M} \\
\bottomrule
\end{tabular}
\end{table}

\subsection{Comparisons with the state-of-the-art}

\textbf{Quantitative comparisons.} 
\tablename~\ref{tab:occ_compare_123} reports the occupancy forecasting results over the next 3 seconds. With direct occupancy input, our method uses explicit motion compensation and sparse delta prediction, achieving the best overall performance with the highest FPS and the fewest parameters. The only exception is the 3s IoU, where our result is slightly lower. For the vision-input setting (-V), we follow $I^2$-World~\cite{I2world} and use the output of the vision-based occupancy prediction model STCOcc~\cite{liao2025stcocc} as the input to the OWM. In this setting, the inference speed is mainly limited by the VON network. Nevertheless, our compact OWM still achieves the best efficiency and performance. \tablename~\ref{tab:occ_compare_456} compares long-term forecasting results under the same setting as GenieDrive~\cite{geniedrive}. Our method shows better long-term consistency. Motion compensation provides a stable prior for future scenes, and the two decoder queries help model scene changes and newly observed regions from historical context. Finally, \tablename~\ref{tab:cross_dataset} shows the direct evaluation results on Occ3D-Waymo under the 2Hz setting. ``Copy-paste'' directly uses the current frame as the future prediction and serves as a lower-bound baseline. Our method achieves the best generalization performance.

\begin{figure}[!t]
\begin{minipage}[c]{0.7\textwidth}
\captionof{table}{Comparison on the long-term occupancy forecasting at 4s, 5s, and 6s in Occ3D-nus.}
\label{tab:occ_compare_456}
\centering
\scriptsize
\vspace{-2mm}
\begin{tabular}{
    r|
    ccc>{\columncolor{gray!20}}c|
    ccc>{\columncolor{gray!20}}c
}
\toprule
    \multirow{2}{*}{Models} & 
    \multicolumn{4}{c|}{mIoU (\%)$\uparrow$} & 
    \multicolumn{4}{c}{IoU (\%)$\uparrow$} \\
    & 
    4s & 5s & 6s & Avg. & 
    4s & 5s & 6s & Avg. \\
\midrule
    OccWorld~\cite{occworld} &
         7.79 &  6.29 &  5.76 &  6.61 &
        17.23 & 15.31 & 14.54 & 15.69 \\
    DOME~\cite{dome} & 
        23.50 & 19.37 & 17.27 & 20.05 & 
        13.84 & 10.14 &  8.15 & 10.71 \\
    UniScene~\cite{uniscene} &
        16.44 & 16.41 & 15.61 & 16.15 & 
        14.31 & 14.33 & 14.15 & 14.26 \\
    $I^2$-World~\cite{I2world} & 
        28.59 & 25.32 & 22.55 & 25.49 &
        41.75 & 38.47 & 35.41 & 38.54 \\
    GenieDrive~\cite{geniedrive} &
        31.16 & 27.17 & 23.66 & 27.33 &
        42.81 & 39.00 & 35.60 & 39.14 \\
    \rowcolor{gray!20}
    \ours (Ours) & 
        \textbf{35.53} & \textbf{31.32} & \textbf{27.42} & \cellcolor{gray!50}\textbf{31.42} &
        \textbf{43.95} & \textbf{40.12} & \textbf{37.79} & \cellcolor{gray!50}\textbf{40.62} \\
\bottomrule
\end{tabular}
\end{minipage}
\hfill
\begin{minipage}[c]{0.28\textwidth}
\vspace{-2mm}
\captionof{table}{Cross-dataset generalization performance in Occ3D-Waymo.}
\label{tab:cross_dataset}
\centering
\scriptsize
\begin{tabular}{
    r|cc
}
\toprule
    Models &
    \makecell[c]{Avg.\\mIoU} &
    \makecell[c]{Avg.\\IoU} \\
\midrule
    Copy\&Paste & 17.17 & 28.21 \\
    $I^2$-World~\cite{I2world} & 36.38 & 52.36 \\
    \rowcolor{gray!20}
    \ours (Ours) & \textbf{38.46} & \textbf{53.19} \\ 
\bottomrule
\end{tabular}
\end{minipage}
\end{figure}

\begin{figure}[!t]
\includegraphics[width=\linewidth]{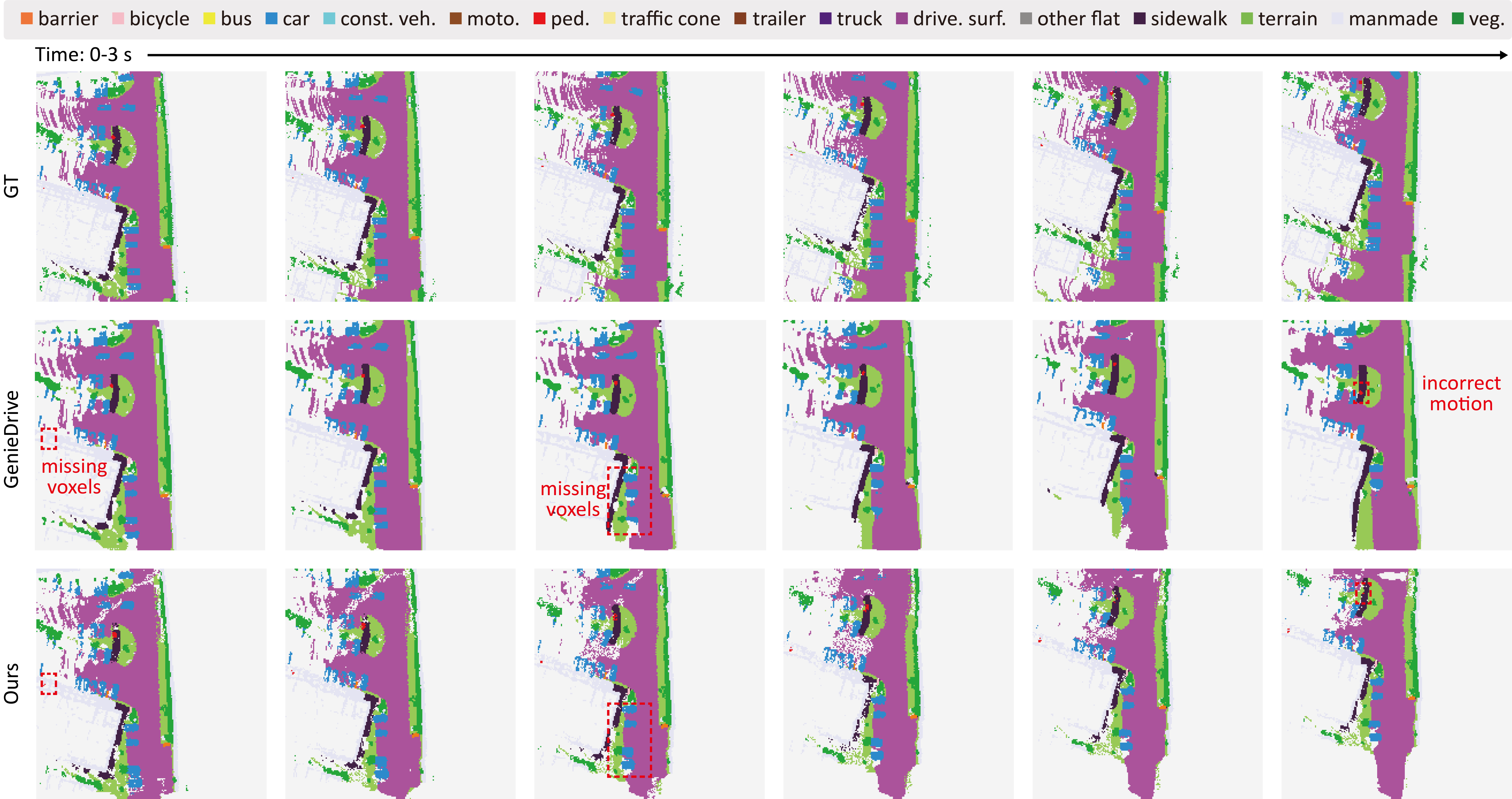}
\vspace{-15pt}
\caption{Occupancy forecasting comparison on the Occ3D-nus val set (zoom in for details).}
\label{fig:vis_compare}
\end{figure}

\textbf{Qualitative comparisons.} 
As shown in \figurename~\ref{fig:vis_compare}, 
GenieDrive misses the pedestrian in the first frame and the vehicle in the third frame, and these errors persist in later predictions. It also predicts the pedestrian motion inaccurately in the last frame. In contrast, our method consistently tracks the pedestrian and keeps the correct number of vehicles in the lower region. For the newly observed region at the bottom, our prediction is less smooth than that of GenieDrive because we do not use a generative diffusion module. Nevertheless, it still recovers a reasonably accurate drivable area.

\begin{wrapfigure}[28]{r}{7.cm}
\captionof{table}{Ablation on key designs in \ours.}
\label{tab:ablation_module}
\centering
\footnotesize
\vspace{-5.3pt}
\begin{tabular}{l|ccc}
\toprule
    \makecell[l]{Idx.\;\;Models} & 
    \makecell[c]{Avg.\\mIoU} & 
    \makecell[c]{Avg.\\IoU} &
    \makecell[c]{FPS}\\
\midrule
    \rowcolor{gray!20}
    \textbf{M0}\;\ours & \textbf{46.87} & \textbf{52.63} & 46 \\
\midrule
    \textbf{M1}\;w/o triplane representation & 41.16 & 47.85 & \textbf{52} \\
    \textbf{M2}\;w/o pretraining triplane & 35.43 & 40.08 & 46 \\ 
    \textbf{M3}\;w/o transformer & 28.69 & 34.36 & 59 \\
    \textbf{M4}\;w/o motion compensation & 31.87 & 37.48 & 43 \\
    \textbf{M5}\;w/o delta prediction & 36.65 & 45.50 & 46 \\
    \textbf{M6}\;w/o finetuning tri. decoder & 45.13 & 51.42 & 46 \\
\bottomrule
\end{tabular}
\\
\captionof{table}{Ablation on hyperparameters.}
\label{tab:ablation_hyper}
\footnotesize
\centering
\begin{tabular}{ccc|ccc}
\toprule
    $G_{\{x,y,z\}}$ & 
    $D_{\rm token}$ & 
    \makecell[c]{depth} & 
    \makecell[c]{Avg.\\mIoU} & 
    \makecell[c]{Avg.\\IoU} &
    \makecell[c]{FPS}\\
\midrule
     8,8,4  &  32 & 1 & 43.94 & 51.77 & \textbf{48} \\
    \rowcolor{gray!20}
     8,8,4  & 128 & 2 & 46.87 & 52.63 & 46 \\
    12,12,3 & 128 & 2 & 46.91 & 53.11 & 43 \\
    12,12,3 & 128 & 8 & 47.05 & 52.49 & 39 \\
    16,16,6 & 256 & 8 & \textbf{49.13} & \textbf{54.89} & 32 \\
\bottomrule
\end{tabular}
\\
\captionof{table}{Ablation on motion compensation.}
\label{tab:ablation_motion_comp}
\footnotesize
\centering
\begin{tabular}{l|ccc}
\toprule
Compensation strategy & 
\makecell[c]{Avg.\\mIoU} & 
\makecell[c]{Avg.\\IoU} &
\makecell[c]{FPS}\\
\midrule
\textbf{C1.1}\;Occ.-space warp & 44.99 & 50.61 & 44 \\
\textbf{C1.2}\;Occ.-space learnable & 43.07 & 48.36 & 42 \\
\rowcolor{gray!20}
\textbf{C2.1}\;Tri.-space warp & 46.87 & 52.63 & 46 \\
\textbf{C2.2}\;Tri.-space learnable & 45.62 & 52.71 & 44 \\
\bottomrule
\end{tabular}
\end{wrapfigure}

\subsection{Ablation study}

\textbf{Effect of key designs in \ours.} 

\tablename~\ref{tab:ablation_module} shows the ablation study of our core modules. M0 is the full model, and the other variants are built by removing key components from M0. M1 removes the $xz$ and $yz$ planes and uses only the $xy$ plane (i.e., BEV). Without height information, it cannot provide accurate 3D predictions. In M2, the triplane and the transformer are trained fully end-to-end. This makes the prediction model rely on unstable triplane features, leading to limited performance. M3 uses only motion compensation without the transformer. Since objects move in the scene, motion compensation causes overlaps. Without the transformer, the model cannot understand scene dynamics or predict future changes, and it cannot infer newly appeared regions, resulting in poor performance. M4 removes motion compensation, and M5 directly predicts the full future scene. Both show clear performance drops. With limited model capacity, misaligned triplane features or direct full-scene prediction makes the learning problem harder. In M6, the triplane decoder is fixed when training the transformer, which leads to a slight performance drop.

\textbf{Effect of hyperparameters.}
\tablename~\ref{tab:ablation_hyper} shows the impact of tokenization grid size $(G_x,G_y,G_z)$, token dimension $D_{\rm token}$, and transformer depth. Larger values improve occupancy forecasting by reducing information loss during tokenization and increasing model capacity, but lower the FPS. The gray row denotes our trade-off version.

\textbf{Effect of motion compensation.}
\tablename~\ref{tab:ablation_motion_comp} compares different motion compensation strategies. All variants use the same model and training settings, and only differ in the compensation strategy. Methods marked with ``warp'' use the explicit compensation operation in Section~3.2. ``Occ.-space'' and ``Tri.-space'' denote the compensation in the occupancy space and the triplane space, respectively. ``learnable'' replaces explicit warping with a learnable alignment module, i.e., Mutual Control Attention in $I^2$-World~\cite{I2world}. C1.1<C2.1 and C1.2<C2.2 show that warping discrete occupancy labels can introduce boundary artifacts, holes, and label discontinuities. Triplane features are smoother and more compact, making alignment more stable. The learnable module adds flexibility, but it leads to less stable performance and lower FPS. We therefore use C2.1 as our trade-off setting.

\begin{figure}[!t]
\begin{minipage}[c]{0.44\textwidth}
\captionof{table}{Ablation on ego and external query counts.}
\label{tab:query_number}
\centering
\small
\begin{tabular}{lcc|cc}
\toprule
    Idx. & 
    \makecell[c]{$N_{\rm ego}$} & 
    \makecell[c]{$N_{\rm ext}$} & 
    \makecell[c]{Avg.\\mIoU}$\uparrow$ & 
    \makecell[c]{Avg.\\IoU}$\uparrow$ \\ 
\midrule
    \textbf{Q1} & 0 & 256 & 40.14 & 50.28 \\
    \rowcolor{gray!20}
    \textbf{Q2} & 64 & 192 & \textbf{46.87} & \textbf{52.63} \\
    \textbf{Q3} & 128 & 128 & 40.53 & 45.35\\
    \textbf{Q4} & 192 & 64 & 41.85 & 50.39 \\
    \textbf{Q5} & 256 & 0 & 41.72 & 46.37\\
\bottomrule
\end{tabular}
\end{minipage}
\hfill
\begin{minipage}[c]{0.5\textwidth}
    \includegraphics[width=\linewidth]{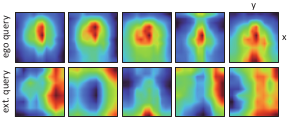}
    \vspace{-15pt}
    \caption{Visualization of ego queries and external queries on all prediction timesteps from Occ3D-nus val set, with five queries per type.}
    \label{fig:query_stat}
\end{minipage}
\vspace{-3mm}
\end{figure}

\textbf{Effect of query counts.}
Ego queries provide explicit control signals that describe future ego motion, while external queries model scene dynamics outside direct control. When either type is insufficient, performance drops. In Q1, without ego queries, the model relies mainly on historical context, leading to weaker controllability. In Q5, without external queries, the model depends mostly on ego motion and cannot fully capture interactions in the scene. Increasing the number of ego queries (from Q1 to Q5) does not consistently improve performance, as they are derived from the same control signal and become redundant. Q2 achieves the best balance between ego guidance and scene modeling. \figurename~\ref{fig:query_stat} shows the statistical behavior of different queries.  For each query, we aggregate its activation over all samples in the Occ3D-nus val set on the $xy$ plane and then normalize it to $[0,1]$. The resulting maps indicate which regions each query mainly contributes to when decoding future state changes. Ego queries exhibit similar patterns and concentrate around the ego vehicle, while external queries cover surrounding areas. This indicates that ego queries mainly encode local control, and a moderate number is sufficient, which explains the strong performance of Q2.

\vspace{-3mm}
\begin{figure}[!h]
\includegraphics[width=\linewidth]{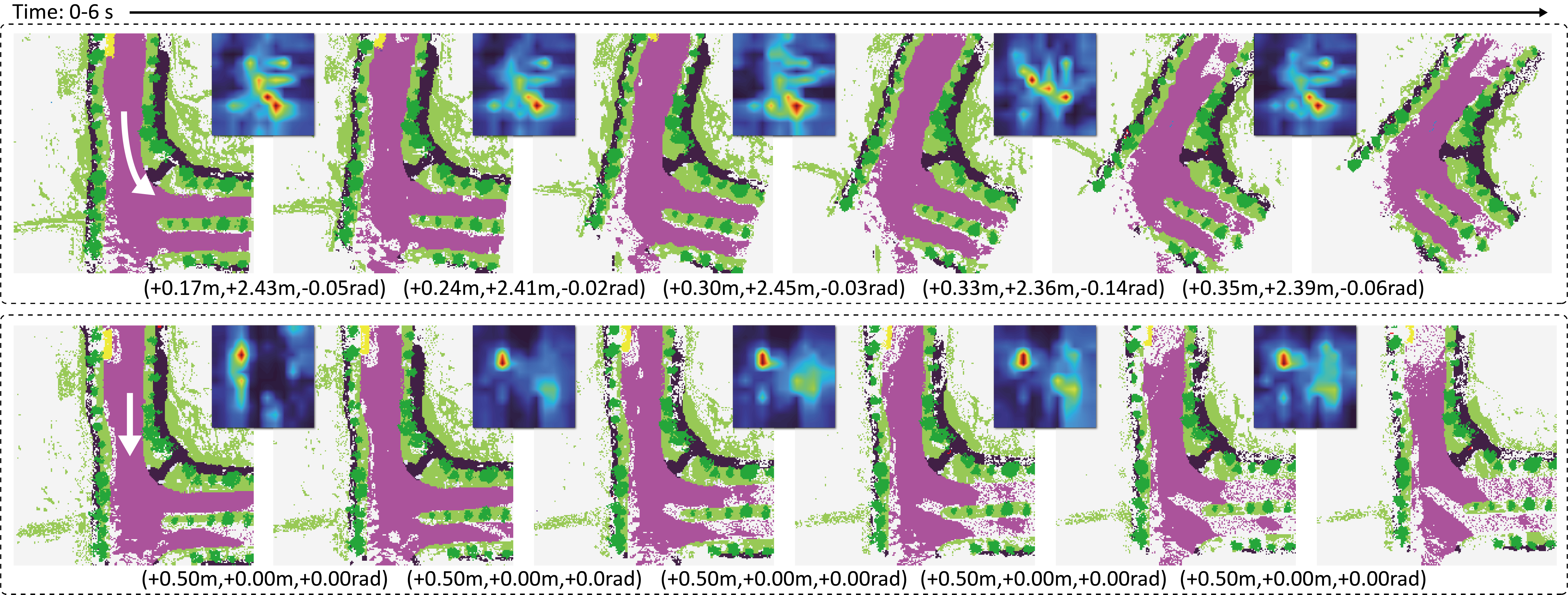}
\vspace{-15pt}
\caption{Controllable future occupancy prediction under different ego-motion inputs.}
\label{fig:ours_control}
\end{figure}
\vspace{-3mm}

\textbf{Effect of controllable ego query.}
The learned external queries are fixed after training, while the ego queries are generated from variable ego-motion inputs, including displacement along the $x$ axis, displacement along the $y$ axis, and the yaw angle in radians. \figurename~\ref{fig:ours_control} shows two examples with the same initial scene but different ego-motion inputs over the next 6 seconds. In each example, adjacent frames show the specific ego-motion control signal and the visualization of one corresponding ego query. The upper example shows the vehicle turning left, while the lower example shows the vehicle going straight. The ego query mainly attends to the region around the ego vehicle and changes consistently with the given control signal, demonstrating the controllability of our method.

\section{Conclusion}
This paper introduces \ours, a compact and efficient 4D OWM that performs both motion compensation and scene-change prediction in the triplane latent space. \ours achieves SOTA performance in short- and long-term forecasting, while delivering faster inference than existing methods.

\begin{ack}
The study was funded by the Shenzhen Science and Technology Program under contracts No. KQTD20240729102051063 and No. JCYJ20241202130025030, the Postdoctoral Fellowship Program (Grade B) of China Postdoctoral Science Foundation under contract No. GZB20260408, the National Natural Science Foundation of China under contracts No. U25B6003, No. 62422602, No. 62372010, No. 62425101, No. 62332002, and the Major Key Project of Pengcheng Laboratory (2024B0101010003). Computing support was provided by Pengcheng Cloudbrain.
\end{ack}

{
\small
\bibliographystyle{ieeenat_fullname}
\bibliography{main}

@inproceedings{occworld,
  title={Occworld: Learning a 3d occupancy world model for autonomous driving},
  author={Zheng, Wenzhao and Chen, Weiliang and Huang, Yuanhui and Zhang, Borui and Duan, Yueqi and Lu, Jiwen},
  booktitle={European conference on computer vision},
  pages={55--72},
  year={2024},
  organization={Springer}
}

@inproceedings{geniedrive,
  title={GenieDrive: Towards Physics-Aware Driving World Model with 4D Occupancy Guided Video Generation},
  author={Yang, Zhenya and Liu, Zhe and Lu, Yuxiang and Hou, Liping and Miao, Chenxuan and Peng, Siyi and Feng, Bailan and Bai, Xiang and Zhao, Hengshuang},
  booktitle={Proceedings of the IEEE/CVF conference on computer vision and pattern recognition},
  year={2026}
}

@inproceedings{i2world,
  title={I2-world: Intra-inter tokenization for efficient dynamic 4d scene forecasting},
  author={Liao, Zhimin and Wei, Ping and Zhang, Ruijie and Chen, Shuaijia and Wang, Haoxuan and Ren, Ziyang},
  booktitle={Proceedings of the IEEE/CVF International Conference on Computer Vision},
  pages={25810--25819},
  year={2025}
}

@article{come,
  title={COME: Adding scene-centric forecasting control to occupancy world model},
  author={Shi, Yining and Jiang, Kun and Meng, Qiang and Wang, Ke and Wang, Jiabao and Sun, Wenchao and Wen, Tuopu and Yang, Mengmeng and Yang, Diange},
  journal={arXiv preprint arXiv:2506.13260},
  year={2025}
}

@inproceedings{wang2024drivedreamer,
  title={Drivedreamer: Towards real-world-drive world models for autonomous driving},
  author={Wang, Xiaofeng and Zhu, Zheng and Huang, Guan and Chen, Xinze and Zhu, Jiagang and Lu, Jiwen},
  booktitle={European conference on computer vision},
  pages={55--72},
  year={2024},
  organization={Springer}
}

@inproceedings{sun2020scalability,
  title={Scalability in perception for autonomous driving: Waymo open dataset},
  author={Sun, Pei and Kretzschmar, Henrik and Dotiwalla, Xerxes and Chouard, Aurelien and Patnaik, Vijaysai and Tsui, Paul and Guo, James and Zhou, Yin and Chai, Yuning and Caine, Benjamin and others},
  booktitle={Proceedings of the IEEE/CVF conference on computer vision and pattern recognition},
  pages={2446--2454},
  year={2020}
}

@inproceedings{uniscene,
  title={Uniscene: Unified occupancy-centric driving scene generation},
  author={Li, Bohan and Guo, Jiazhe and Liu, Hongsi and Zou, Yingshuang and Ding, Yikang and Chen, Xiwu and Zhu, Hu and Tan, Feiyang and Zhang, Chi and Wang, Tiancai and others},
  booktitle={Proceedings of the computer vision and pattern recognition conference},
  pages={11971--11981},
  year={2025}
}

@article{occsora,
  title={Occsora: 4d occupancy generation models as world simulators for autonomous driving},
  author={Wang, Lening and Zheng, Wenzhao and Ren, Yilong and Jiang, Han and Cui, Zhiyong and Yu, Haiyang and Lu, Jiwen},
  journal={arXiv preprint arXiv:2405.20337},
  year={2024}
}

@article{occllama,
  title={Occllama: An occupancy-language-action generative world model for autonomous driving},
  author={Wei, Julong and Yuan, Shanshuai and Li, Pengfei and Hu, Qingda and Gan, Zhongxue and Ding, Wenchao},
  journal={arXiv preprint arXiv:2409.03272},
  year={2024}
}

@inproceedings{occllm,
  title={Occ-LLM: Enhancing Autonomous Driving with Occupancy-Based Large Language Models},
  author={Xu, Tianshuo and Lu, Hao and Yan, Xu and Cai, Yingjie and Liu, Bingbing and Chen, Yingcong},
  booktitle={International Conference on Robotics and Automation},
  year={2025}
}

@inproceedings{renderworld,
  title={Renderworld: World model with self-supervised 3d label},
  author={Yan, Ziyang and Dong, Wenzhen and Shao, Yihua and Lu, Yuhang and Haiyang, Liu and Liu, Jingwen and Wang, Haozhe and Wang, Zhe and Wang, Yan and Remondino, Fabio and others},
  booktitle={International Conference on Robotics and Automation},
  year={2025}
}

@article{dome,
  title={Dome: Taming diffusion model into high-fidelity controllable occupancy world model},
  author={Gu, Songen and Yin, Wei and Jin, Bu and Guo, Xiaoyang and Wang, Junming and Li, Haodong and Zhang, Qian and Long, Xiaoxiao},
  journal={arXiv preprint arXiv:2410.10429},
  year={2024}
}

@inproceedings{vad,
  title={Vad: Vectorized scene representation for efficient autonomous driving},
  author={Jiang, Bo and Chen, Shaoyu and Xu, Qing and Liao, Bencheng and Chen, Jiajie and Zhou, Helong and Zhang, Qian and Liu, Wenyu and Huang, Chang and Wang, Xinggang},
  booktitle={Proceedings of the IEEE/CVF International Conference on Computer Vision},
  pages={8340--8350},
  year={2023}
}

@article{ha2018world,
  title={World models},
  author={Ha, David and Schmidhuber, J{\"u}rgen},
  journal={arXiv preprint arXiv:1803.10122},
  year={2018}
}

@article{ha2018recurrent,
  title={Recurrent world models facilitate policy evolution},
  author={Ha, David and Schmidhuber, J{\"u}rgen},
  journal={Advances in neural information processing systems},
  volume={31},
  year={2018}
}

@inproceedings{liu2024volumetric,
  title={Volumetric environment representation for vision-language navigation},
  author={Liu, Rui and Wang, Wenguan and Yang, Yi},
  booktitle={Proceedings of the IEEE/CVF Conference on Computer Vision and Pattern Recognition},
  pages={16317--16328},
  year={2024}
}

@inproceedings{xia2023scpnet,
  title={Scpnet: Semantic scene completion on point cloud},
  author={Xia, Zhaoyang and Liu, Youquan and Li, Xin and Zhu, Xinge and Ma, Yuexin and Li, Yikang and Hou, Yuenan and Qiao, Yu},
  booktitle={Proceedings of the IEEE/CVF conference on computer vision and pattern recognition},
  pages={17642--17651},
  year={2023}
}

@inproceedings{cao2024pasco,
  title={Pasco: Urban 3d panoptic scene completion with uncertainty awareness},
  author={Cao, Anh-Quan and Dai, Angela and De Charette, Raoul},
  booktitle={Proceedings of the IEEE/CVF Conference on Computer Vision and Pattern Recognition},
  pages={14554--14564},
  year={2024}
}

@inproceedings{marinello2025camera,
  title={Camera-Only 3D Panoptic Scene Completion for Autonomous Driving through Differentiable Object Shapes},
  author={Marinello, Nicola and Cassiman, Simen and Heylen, Jonas and Proesmans, Marc and Van Gool, Luc},
  booktitle={Proceedings of the Computer Vision and Pattern Recognition Conference},
  pages={2520--2529},
  year={2025}
}

@inproceedings{song2017semantic,
  title={Semantic scene completion from a single depth image},
  author={Song, Shuran and Yu, Fisher and Zeng, Andy and Chang, Angel X and Savva, Manolis and Funkhouser, Thomas},
  booktitle={Proceedings of the IEEE conference on computer vision and pattern recognition},
  pages={1746--1754},
  year={2017}
}

@inproceedings{li2023voxformer,
  title={Voxformer: Sparse voxel transformer for camera-based 3d semantic scene completion},
  author={Li, Yiming and Yu, Zhiding and Choy, Christopher and Xiao, Chaowei and Alvarez, Jose M and Fidler, Sanja and Feng, Chen and Anandkumar, Anima},
  booktitle={Proceedings of the IEEE/CVF conference on computer vision and pattern recognition},
  pages={9087--9098},
  year={2023}
}

@inproceedings{viewformer,
  title={Viewformer: Exploring spatiotemporal modeling for multi-view 3d occupancy perception via view-guided transformers},
  author={Li, Jinke and He, Xiao and Zhou, Chonghua and Cheng, Xiaoqiang and Wen, Yang and Zhang, Dan},
  booktitle={European Conference on Computer Vision},
  pages={90--106},
  year={2024},
  organization={Springer}
}

@inproceedings{occnet,
  title={Scene as occupancy},
  author={Tong, Wenwen and Sima, Chonghao and Wang, Tai and Chen, Li and Wu, Silei and Deng, Hanming and Gu, Yi and Lu, Lewei and Luo, Ping and Lin, Dahua and others},
  booktitle={Proceedings of the IEEE/CVF International Conference on Computer Vision},
  pages={8406--8415},
  year={2023}
}

@article{wu2024holodrive,
  title={HoloDrive: Holistic 2D-3D Multi-Modal Street Scene Generation for Autonomous Driving},
  author={Wu, Zehuan and Ni, Jingcheng and Wang, Xiaodong and Guo, Yuxin and Chen, Rui and Lu, Lewei and Dai, Jifeng and Xiong, Yuwen},
  journal={arXiv preprint arXiv:2412.01407},
  year={2024}
}

@inproceedings{zhou2017unsupervised,
  title={Unsupervised learning of depth and ego-motion from video},
  author={Zhou, Tinghui and Brown, Matthew and Snavely, Noah and Lowe, David G},
  booktitle={Proceedings of the IEEE conference on computer vision and pattern recognition},
  pages={1851--1858},
  year={2017}
}

@article{jaderberg2015spatial,
  title={Spatial transformer networks},
  author={Jaderberg, Max and Simonyan, Karen and Zisserman, Andrew and others},
  journal={Advances in neural information processing systems},
  volume={28},
  year={2015}
}

@inproceedings{huang2023tri,
  title={Tri-perspective view for vision-based 3d semantic occupancy prediction},
  author={Huang, Yuanhui and Zheng, Wenzhao and Zhang, Yunpeng and Zhou, Jie and Lu, Jiwen},
  booktitle={Proceedings of the IEEE/CVF conference on computer vision and pattern recognition},
  pages={9223--9232},
  year={2023}
}

@article{jia2023adriver,
  title={Adriver-i: A general world model for autonomous driving},
  author={Jia, Fan and Mao, Weixin and Liu, Yingfei and Zhao, Yucheng and Wen, Yuqing and Zhang, Chi and Zhang, Xiangyu and Wang, Tiancai},
  journal={arXiv preprint arXiv:2311.13549},
  year={2023}
}

@inproceedings{liao2025stcocc,
  title={Stcocc: Sparse spatial-temporal cascade renovation for 3d occupancy and scene flow prediction},
  author={Liao, Zhimin and Wei, Ping and Chen, Shuaijia and Wang, Haoxuan and Ren, Ziyang},
  booktitle={Proceedings of the Computer Vision and Pattern Recognition Conference},
  pages={1516--1526},
  year={2025}
}

@inproceedings{li2025end,
  title={End-to-end driving with online trajectory evaluation via bev world model},
  author={Li, Yingyan and Wang, Yuqi and Liu, Yang and He, Jiawei and Fan, Lue and Zhang, Zhaoxiang},
  booktitle={Proceedings of the IEEE/CVF International Conference on Computer Vision},
  pages={27137--27146},
  year={2025}
}

@article{zhang2024bevworld,
  title={Bevworld: A multimodal world model for autonomous driving via unified bev latent space},
  author={Zhang, Yumeng and Gong, Shi and Xiong, Kaixin and Ye, Xiaoqing and Tan, Xiao and Wang, Fan and Huang, Jizhou and Wu, Hua and Wang, Haifeng},
  journal={arXiv preprint arXiv:2407.05679},
  year={2024}
}

@inproceedings{wang2024driving,
  title={Driving into the future: Multiview visual forecasting and planning with world model for autonomous driving},
  author={Wang, Yuqi and He, Jiawei and Fan, Lue and Li, Hongxin and Chen, Yuntao and Zhang, Zhaoxiang},
  booktitle={Proceedings of the IEEE/CVF Conference on Computer Vision and Pattern Recognition},
  pages={14749--14759},
  year={2024}
}

@article{gao2025vista,
  title={Vista: A generalizable driving world model with high fidelity and versatile controllability},
  author={Gao, Shenyuan and Yang, Jiazhi and Chen, Li and Chitta, Kashyap and Qiu, Yihang and Geiger, Andreas and Zhang, Jun and Li, Hongyang},
  journal={Advances in Neural Information Processing Systems},
  volume={37},
  pages={91560--91596},
  year={2025}
}

@inproceedings{surroundocc,
  title={Surroundocc: Multi-camera 3d occupancy prediction for autonomous driving},
  author={Wei, Yi and Zhao, Linqing and Zheng, Wenzhao and Zhu, Zheng and Zhou, Jie and Lu, Jiwen},
  booktitle={Proceedings of the IEEE/CVF International Conference on Computer Vision},
  pages={21729--21740},
  year={2023}
}

@article{occ3d,
  title={Occ3d: A large-scale 3d occupancy prediction benchmark for autonomous driving},
  author={Tian, Xiaoyu and Jiang, Tao and Yun, Longfei and Mao, Yucheng and Yang, Huitong and Wang, Yue and Wang, Yilun and Zhao, Hang},
  journal={Advances in Neural Information Processing Systems},
  volume={36},
  pages={64318--64330},
  year={2023}
}

@article{lu2021monet,
  title={Monet: Motion-based point cloud prediction network},
  author={Lu, Fan and Chen, Guang and Li, Zhijun and Zhang, Lijun and Liu, Yinlong and Qu, Sanqing and Knoll, Alois},
  journal={IEEE Transactions on Intelligent Transportation Systems},
  volume={23},
  number={8},
  pages={13794--13804},
  year={2021},
  publisher={IEEE}
}

@inproceedings{carion2020end,
  title={End-to-end object detection with transformers},
  author={Carion, Nicolas and Massa, Francisco and Synnaeve, Gabriel and Usunier, Nicolas and Kirillov, Alexander and Zagoruyko, Sergey},
  booktitle={European conference on computer vision},
  pages={213--229},
  year={2020},
  organization={Springer}
}

@article{vaswani2017attention,
  title={Attention is all you need},
  author={Vaswani, Ashish and Shazeer, Noam and Parmar, Niki and Uszkoreit, Jakob and Jones, Llion and Gomez, Aidan N and Kaiser, {\L}ukasz and Polosukhin, Illia},
  journal={Advances in neural information processing systems},
  volume={30},
  year={2017}
}

@inproceedings{mersch2022self,
  title={Self-supervised point cloud prediction using 3d spatio-temporal convolutional networks},
  author={Mersch, Benedikt and Chen, Xieyuanli and Behley, Jens and Stachniss, Cyrill},
  booktitle={Conference on Robot Learning},
  pages={1444--1454},
  year={2022},
  organization={PMLR}
}

@inproceedings{khurana2023point,
  title={Point cloud forecasting as a proxy for 4d occupancy forecasting},
  author={Khurana, Tarasha and Hu, Peiyun and Held, David and Ramanan, Deva},
  booktitle={Proceedings of the IEEE/CVF Conference on Computer Vision and Pattern Recognition},
  pages={1116--1124},
  year={2023}
}

@inproceedings{cam4docc,
  title={Cam4docc: Benchmark for camera-only 4d occupancy forecasting in autonomous driving applications},
  author={Ma, Junyi and Chen, Xieyuanli and Huang, Jiawei and Xu, Jingyi and Luo, Zhen and Xu, Jintao and Gu, Weihao and Ai, Rui and Wang, Hesheng},
  booktitle={Proceedings of the IEEE/CVF Conference on Computer Vision and Pattern Recognition},
  pages={21486--21495},
  year={2024}
}

@inproceedings{occprophet,
  title={OccProphet: Pushing Efficiency Frontier of Camera-Only 4D Occupancy Forecasting with Observer-Forecaster-Refiner Framework},
  author={Chen, Junliang and Xu, Huaiyuan and Wang, Yi and Chau, Lap-Pui},
  booktitle={ICLR},
  year={2025}
}

@article{xu2024spatiotemporal,
  title={Spatiotemporal Decoupling for Efficient Vision-Based Occupancy Forecasting},
  author={Xu, Jingyi and Chen, Xieyuanli and Ma, Junyi and Huang, Jiawei and Xu, Jintao and Wang, Yue and Pei, Ling},
  journal={arXiv preprint arXiv:2411.14169},
  year={2024}
}

@inproceedings{cmop,
  title={Exploiting Continuous Motion Clues for Vision-Based Occupancy Prediction},
  author={Xu, Haoran and Peng, Peixi and Zhang, Xinyi and Tan, Guang and Li, Yaokun and Wang, Shuaixian and Li, Luntong},
  booktitle={AAAI},
  year={2025}
}

@inproceedings{semcity,
  title={Semcity: Semantic scene generation with triplane diffusion},
  author={Lee, Jumin and Lee, Sebin and Jo, Changho and Im, Woobin and Seon, Juhyeong and Yoon, Sung-Eui},
  booktitle={Proceedings of the IEEE/CVF conference on computer vision and pattern recognition},
  pages={28337--28347},
  year={2024}
}

@inproceedings{berman2018lovasz,
  title={The lov{\'a}sz-softmax loss: A tractable surrogate for the optimization of the intersection-over-union measure in neural networks},
  author={Berman, Maxim and Triki, Amal Rannen and Blaschko, Matthew B},
  booktitle={Proceedings of the IEEE conference on computer vision and pattern recognition},
  pages={4413--4421},
  year={2018}
}

@inproceedings{nuscenes,
  title={nuscenes: A multimodal dataset for autonomous driving},
  author={Caesar, Holger and Bankiti, Varun and Lang, Alex H and Vora, Sourabh and Liong, Venice Erin and Xu, Qiang and Krishnan, Anush and Pan, Yu and Baldan, Giancarlo and Beijbom, Oscar},
  booktitle={Proceedings of the IEEE/CVF conference on computer vision and pattern recognition},
  pages={11621--11631},
  year={2020}
}

@inproceedings{cao2022monoscene,
  title={Monoscene: Monocular 3d semantic scene completion},
  author={Cao, Anh-Quan and De Charette, Raoul},
  booktitle={Proceedings of the IEEE/CVF Conference on Computer Vision and Pattern Recognition},
  pages={3991--4001},
  year={2022}
}
}

\end{document}